# Behavior and performance of the deep belief networks on image classification

**Karol Gregor,**
New York University, New York, NY, 10003 &
California Institute of Technology, Pasadena, CA, 91125
kgregor@cs.nyu.edu

**Gregory Griffin,**
California Institute of Technology
Pasadena, CA, 91125
griffin@caltech.edu

## Abstract

We apply deep belief networks of restricted Boltzmann machines to bags of words of sift features obtained from databases of 13 Scenes, 15 Scenes and Caltech 256 and study experimentally their behavior and performance. We find that the final performance in the supervised phase is reached much faster if the system is pre-trained. Pre-training the system on a larger dataset keeping the supervised dataset fixed improves the performance (for the 13 Scenes case). After the unsupervised pre-training, neurons arise that form approximate explicit representations for several categories (meaning they are mostly active for this category). The last three facts suggest that unsupervised training really discovers structure in these data. Pre-training can be done on a completely different dataset (we use Corel dataset) and we find that the supervised phase performs just as good (on the 15 Scenes dataset). This leads us to conjecture that one can pre-train the system once (e.g. in a factory) and subsequently apply it to many supervised problems which then learn much faster. The best performance is obtained with single hidden layer system suggesting that the histogram of sift features doesn't have much high level structure. The overall performance is almost equal, but slightly worse then that of the support vector machine and the spatial pyramidal matching.

## 1 Introduction

Deep belief networks (DBN) of restricted Boltzmann machines (RBM) [1], along with the back-propagation algorithm and recently discovered layer by layer unsupervised learning algorithm of contrastive divergence [2] have recently become a popular topic of research in supervised and unsupervised classification because they promise several advantages. They have fast inference, fast unsupervised learning, are rather flexible, and are deep. One last advantage is the ability to encode richer, higher order structures [3] as in the hierarchical structure of the mammalian cerebral cortex.

The goal of this paper is to study the applicability of the deep belief networks to visual image classification. To this effect we use popular image datasets: 13 Scenes dataset, a newer 15 Scenes dataset, Caltech 256 and Corel dataset. The DBN's are not applied directly to images. Instead, they are applied to histograms of frequencies of visual words, which are obtained by vector quantization of SIFT features [4, 5]. More generally the image is broken up into 2x2 or 4x4 regions, treating the words in each regions as different (with some smoothing function) to capture some spatial information and making histogram of these. The code used for implementing and testing hierarchical restricted Boltzmann machines is that on the Geoffrey Hinton's website with appropriate modifications.

The architecture and the training are those described in [2]. The values in the input layer are real valued normalized histograms of the visual words [4]. These are connected to 0,1,2 or 3 hidden layers. Finally the top layer containing the labels is added. The network is first trained in an unsupervised way (except for the 0 layer system), which is termed pre-training, using contrastive divergence. Subsequently the layer of labels is added and the network is trained using back-propagation.

We look at the 13 Scenes, 15 Scenes and Caltech 256 databases in turn, analyzing most extensively the 13 Scenes database. Apart from dependencies on few parameters and overall performance we are especially interested in understanding the effect of the unsupervised phase. We discuss how fast the network learns during the supervised phase compared to case without pre-training, whether it can take advantage of a large amount of unsupervised data, whether these data can be of different nature then the labeled data and what do some of the actual neurons represent.

## 2 13 Scenes

In this section we use the database of 3859 images of 13 Scenes [5]. Every image is first processed in the following way: One extracts patches of some size N using a regular grid of spacing l. One then sub-samples the image by a factor of two and repeats. This is repeated until the image is small enough that only one patch can be extracted. The resulting patches are transformed into sift descriptors and are clustered using K-means algorithm into 1000 visual words. One more word is added that specifies how many of the patches have overall variation below certain threshold and which were excluded from the above clustering. Thus we start we a bag of 1001 words. The architecture and training is as described in the introduction, with the layer of labels containing 13 neurons.

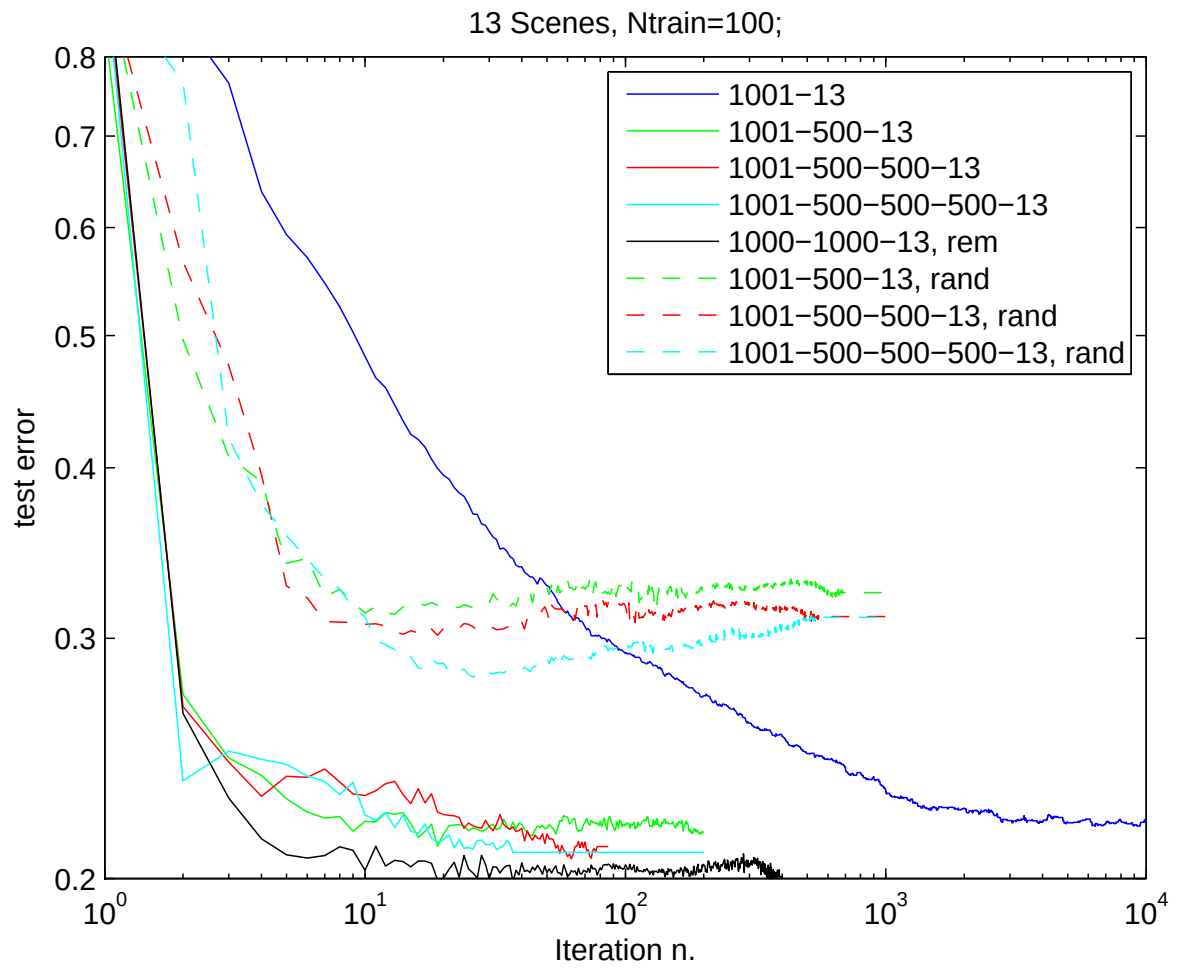

Figure 1: Performance on the 13 Scenes database. The notation e.g. 1001-500-13 means: input of 1001 units - hidden layer 1 of 500 units - output of 13 units. The networks containing hidden layers were trained layer by layer using contrastive divergence for 800 iterations (sweeps through the training set). Then the networks were trained using back-propagation. The final performances between networks are similar, but pre-trained networks have much shorter supervised phase. Word rem means that the special visual word was removed. Also shown are the cases of hidden layer systems with no pre-training but starting with a random small initial weights (rand).

The loglog plot of the dependence of performance on the number of iterations (number of sweeps through training set) for various architectures is in the Fig. 1. For the network with zero hidden layers (1001-13) there is no unsupervised phase and for the networks with positive number of hidden layers, each unsupervised pre-training consisted of 800 sweeps through the entire sample. The dashed lines show the performance of networks with hidden units without pre-training and starting with small random initial conditions.

First, we observe that the final performance is better for the deep network then for the zero hidden layer one, but only slightly. Next, we observe that once the system with hidden layers is pre-trained in the unsupervised way, it is much faster to obtain a good performance. We see that after a single iteration through the entire sample one gets around 25% error, something which took more then 500 sweeps to achieve with the zero layer network.

Let us look at the actual run times. The zero layer saturates after about 1000-2000 iterations. 1000 iterations took about 43 minutes to run. In the 1001-500-13 network, to get as good performance we need about 200 pre-training periods (see below). This took about 28 minutes. For the back-prop we need about 10 iterations which took about 6 minutes (full back-prop for single hidden layer). So single hidden layer system took about 34 minutes to train.

Let us note there there are two basic ways of running the back-propagation. Either on the entire network or only on the top layer. The later is much faster (as fast per iteration as the zero layer system if the number of units in the final layer is the same as in the input). We find that for the system with one hidden layer, the performance of both methods is roughly the same as the function of iteration but for more hidden layer, the performance of the later gets worse.

The network has a large number of parameters and overtrains very quickly. It reaches 100% performance on the train set after 25, 10, 8 iterations for the network with 1,2,3 hidden layers respectively. Despite this, it performs well on the test set, and the test error does not start to increase in time (probably due to a weight decay in the algorithm).

Next it is interesting to look at the performance as a function of the number of pre-training iterations. This is shown in the Fig. 1. We see that increasing this number helps, up to a certain point.

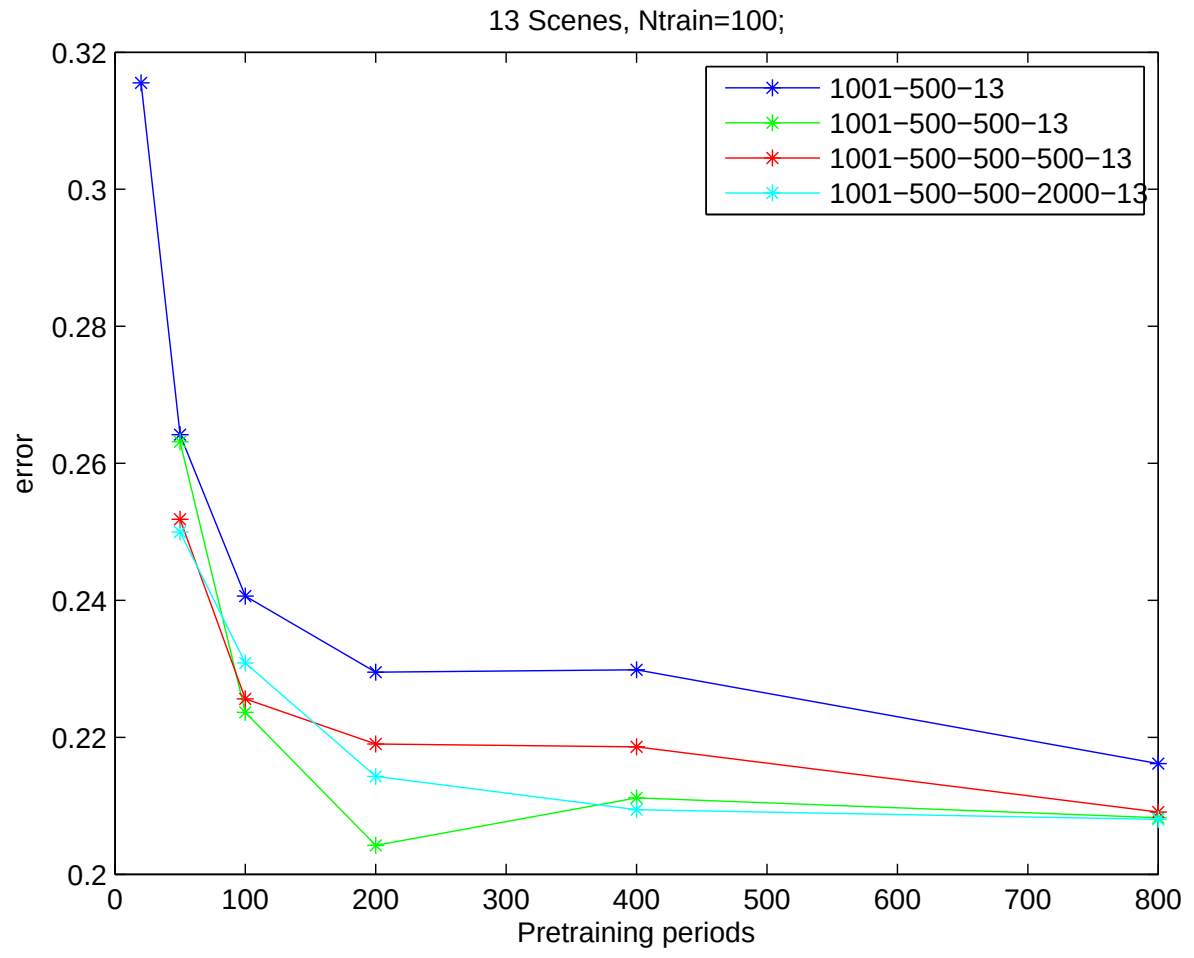

Figure 2: Performance as a function of the number of the pre-training epochs. Increasing the later first improves the performance and then saturates.

We want a good learning system to learn about the world in an unsupervised way by discovering its structure and we should provide supervision only part of the time. To test this principle, we pre-train the network on 100 examples from each category for 800 iterations. Then we run the supervised phase with a subset of this training set containing 1,2,4,8,16,32,64 examples and test on the testing set of the images not contained in the 100 examples per category used for pre-training. The results are shown in the Fig. 3 (along other Scenes results). We see that indeed we get the better performance in this case, though the improvement is not impressive.

In the brain the implicit information in the image is transformed into and explicit representation the in the higher level brain, meaning that, speaking in idealized terms, there are neurons which fire if and only if a given concept was present. In our example, this would correspond to finding neurons that would be active if and only a picture from a given category is present. We now look into whether or how much is this true.

We train the network in an unsupervised way as before. We consider a given layer, namely layer three. We choose a category. We present the images one by one to the network and look at the activities of each neuron. For every neuron we look at how the activity of that neuron is able to distinguish a chosen category from the other categories. We look at the following quantity. Given a threshold - a real number between zero and one. For all the images from the chosen category we calculate the percentage of cases it classifies as being from that category. For all the images not from the chosen category we calculate the percentage it classifies as not being from that category. We average the two. Then we maximize over the thresholds. We call this the performance parameter. Then we chose the neuron with the best performance parameter. We do this for each category. The results are in the Fig 4, with the description in the caption. Actually for few categories we picked the neurons by hand, if the activities looked seemingly sharper, but the results are similar.

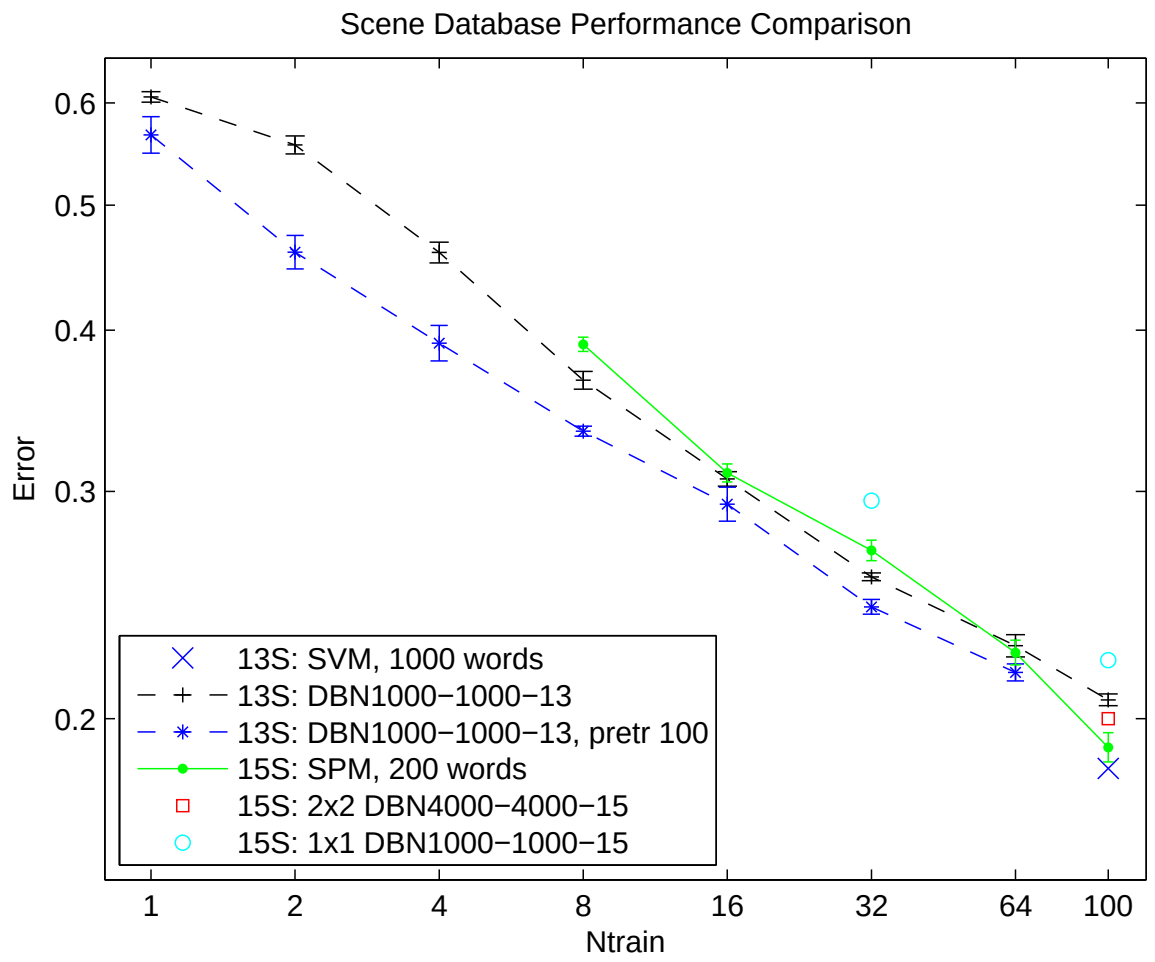

Figure 3: Results for 13 and 15 Scenes databases. Notation: 13S (13 Scenes database), 15S (15 Scenes database), DBN (Deep belief network), SVM (Support vector machine), SPM (Spatial pyramidal matching), pretr 100 - network was pre-trained on 100 examples per category. The overall performance of DBN was close to other methods, just a but little lower. Notice that the pre-training on 100 examples improved the performance.

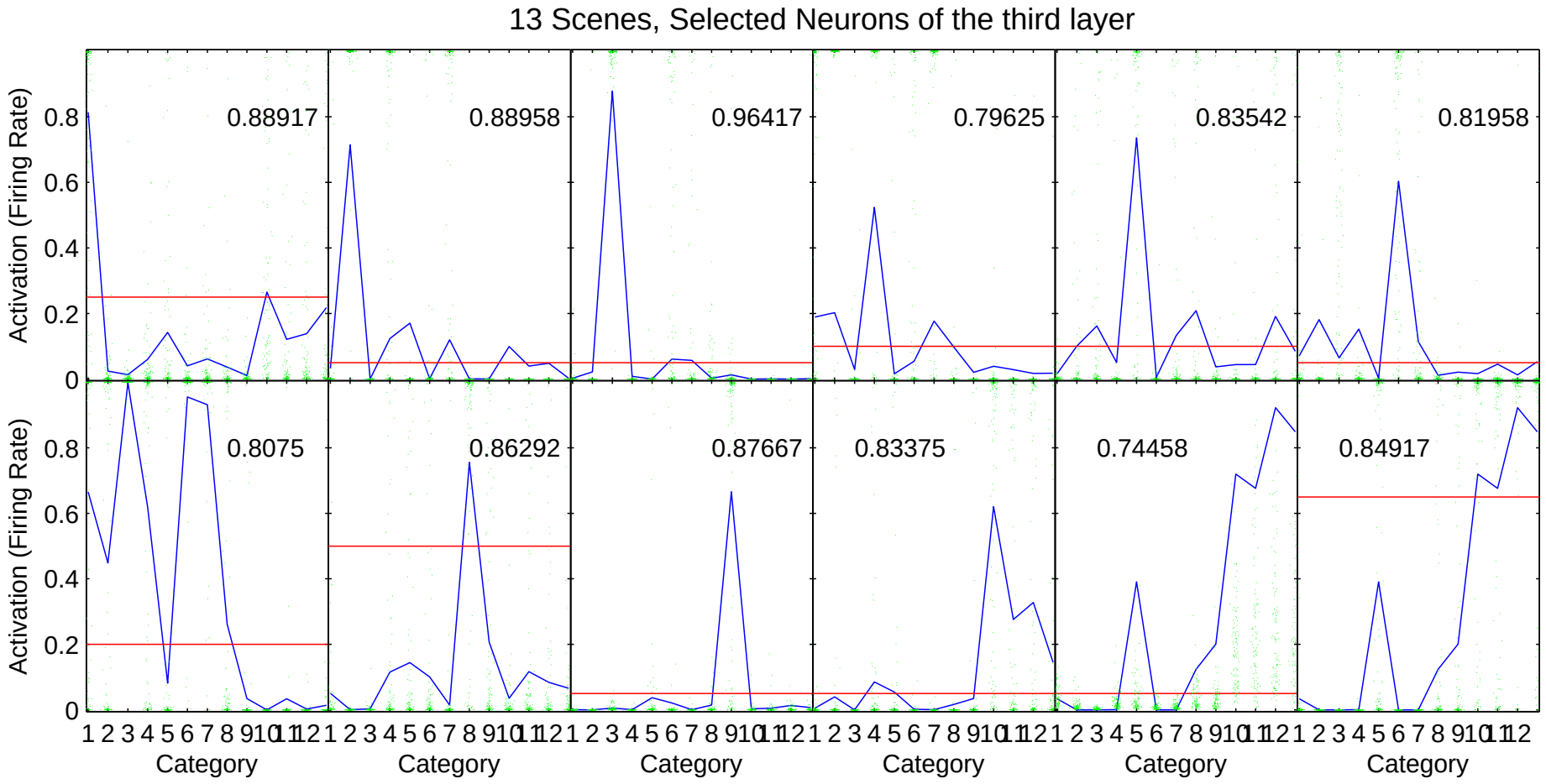

Figure 4: Neurons closest to explicitly representing a given category. Each pair of axes shows activity of a particular neuron. They were chosen by hand (but close to the prescription in the text) to best represent each category (in the order from left to right, from top to bottom). The blue curve is the average activity over all images in a given category (of x-axis). The number is the performance parameter as defined in the text. The red line is the best threshold, as defined in the text. Each green dot is the activity for a particular image (spread randomly in the interval (-0.1,0.1) around each point to give the idea of a density). The neuron for category 13 is not shown.

We see that neurons $1, 2, 3, 5, 8, 9$ are relatively nice explicit representations. Neuron $7$ is not able to distinguish categories but distinguish three categories from the rest. Similarly the neurons $10$,$11$,$12$,$13$ are not able to distinguish one another but distinguish themselves from the rest. This is reasonable since they are all indoor rooms and are more similar to each other then to various outdoor scenes. The best activities at other layers were somewhat similar, but the third layer happens to give the best results.

A priori the network is symmetric under flipping the activity of a neuron when we flip appropriate connections. For this reason we flipped neurons with average activity more then $0.5$. However the problem is not fully symmetric due to weight decay and initial conditions and the actual percentage of such neurons as a function of layer number is $48.2\%$, $7.4\%$, $0.4\%$, $0\%$ and $0\%$.

The explicit representations emerging in the network are nice only if the same cannot be said about looking at the data itself. Now the data are small ($\sim 0.001$) real numbers so we cannot be looking at average activities because they can be swamped by few images having large values. However we can look at the threshold defined above which lets us distinguish categories. The comparison is shown in the Figure 5. We see that the performance as measured by this parameter is somewhat better. In either case, it is nice that there are neurons from which we can read off if the picture belongs to a given category or not by just looking at the overall activity. One also naturally asks, since we are picking the best neurons, whether random connections wouldn't also produce something that looks like the above graph. We checked that this is not so and the graphs are much more uniform.

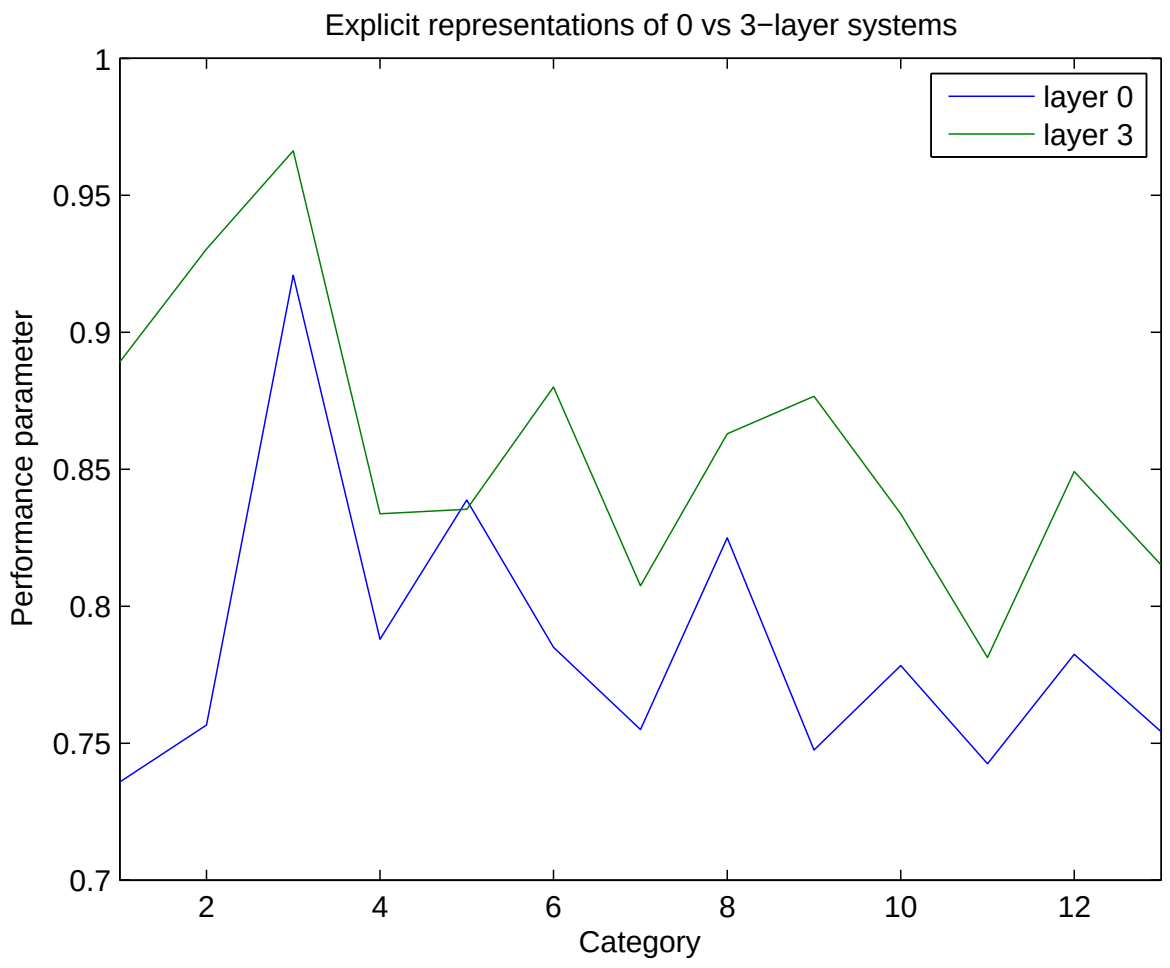

Figure 5: The performance parameter as defined in the text. The third layer neurons distinguish categories better then the input units do.

Figure 3 shows comparison between the performance obtained here and support vector machine (SVM). We see that DBN performance is close but not quite equal to this. The 15 Scenes results are explained in the next section.

## 3 15 Scenes

The data set in this section is the same as above with two scenes added. There are only historical reasons for having these two. There is also some difference in extracting the visual words. The patches were chosen using a different grid and they were clustered using mixture of Gaussians rather then K-means. We study the system for 200 and 1000 visual words (clusters). We also break up the image into 4 regions on 2x2 grid and treating word in each region as different (we also use some smoothing function). The results are shown in the Figure 6.

We see that the 1000 words cases do better then the 200 ones and the 2x2 grid case does better then the rest. For simple bag of words, the one hidden layer, 1000-1000-15 system, does the best and adding more layers worsens the performance. Again we see that after pre-training, as in the 13

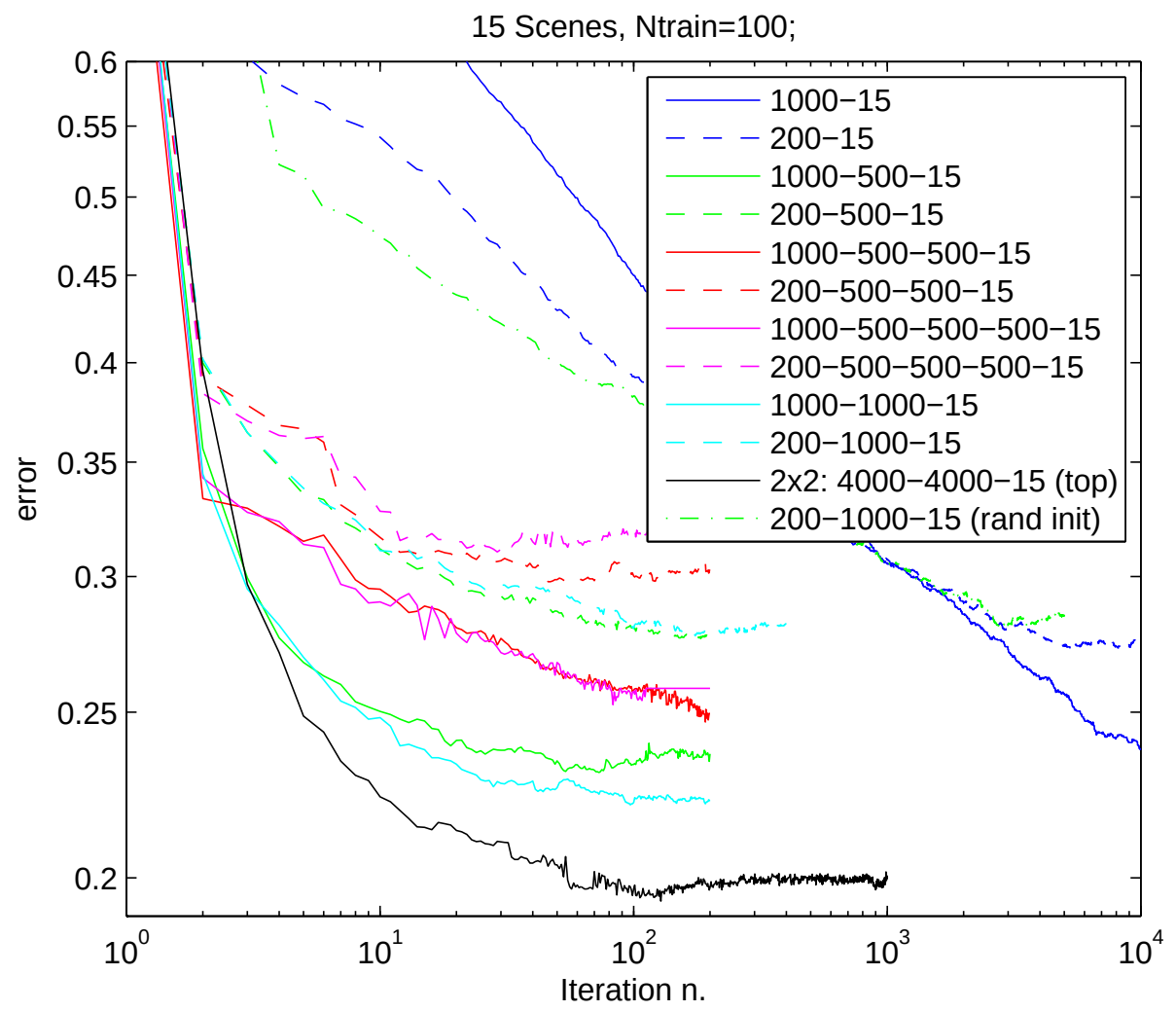

Figure 6: The performance on the 15 scenes database. For description see the caption of Fig 1

Scenes case, a good performance is reached after much fewer iterations than without pre-training, though longer then in that case.

Next we compare the performance of the system for various methods and various sizes of the training set. We calculate the performance for the cases when the same training set is used for both pre-training and back-prop. Then for the case where pre-training is done on 100 images per category and back-prop for smaller number of them. Then when we use back-prop only, starting with small initial conditions. Finally we test whether network can learn in an unsupervised way from one data-set and then train in a supervised way using another data-set. For this purpose, Corel dataset and 15 Scenes dataset were used respectively.

The results for these four ways are shown in the Fig. 7. We see that in this case, no matter which method is used, the final result is the same. For the performance as a function of the number of iterations all the methods with pre-training behaved roughly the same way. The case without pre-training was much slower. To find a case in between, we also used $1\%$ of Corel images and in this case the performance increased a bit more slowly then in the previous pre-trained cases. The final results for second and third layer are similar, but slightly worse. The Corel pre-training is particularly interesting because it suggests that we can pre-train the system once on one data-set, e.g. one can buy a factory pre-trained product, and then one can use fast supervised training for various different problems. However, more tests are needed to establish this conjecture.

It is not quite clear why for the 15 Scenes, as opposed to the 13 Scenes, all these ways of running the problem give the same final performance. One reason might be that these experiments are done on 200 words instead of the 1000. Other might be that somehow there is not as much structure in these data for unsupervised training to discover and one really needs the labels to separate the data.

Finally we compare the performance to the spatial pyramidal matching (SPM) [6], Fig 3. The SPM takes advantage of a spatial information. We see it performs better then simple bag of words, but roughly the same as 2x2 system. On the other hand the 2x2 system used 1000 visual words (quantization points) and SPM only 200. It would be interesting to see whether 1000 words would push the performance higher, considering the fact that increasing the number of words for DBN helps.

## 4 Caltech 256

We performed the basic analysis also on Caltech 256. The features were extracted the same way as for the 15 Scene database. We applied applied the RBM's to bag of 200 words and also, bag of 3200 words obtained by breaking the region up into 4x4 grid. To save space, we only show the later,

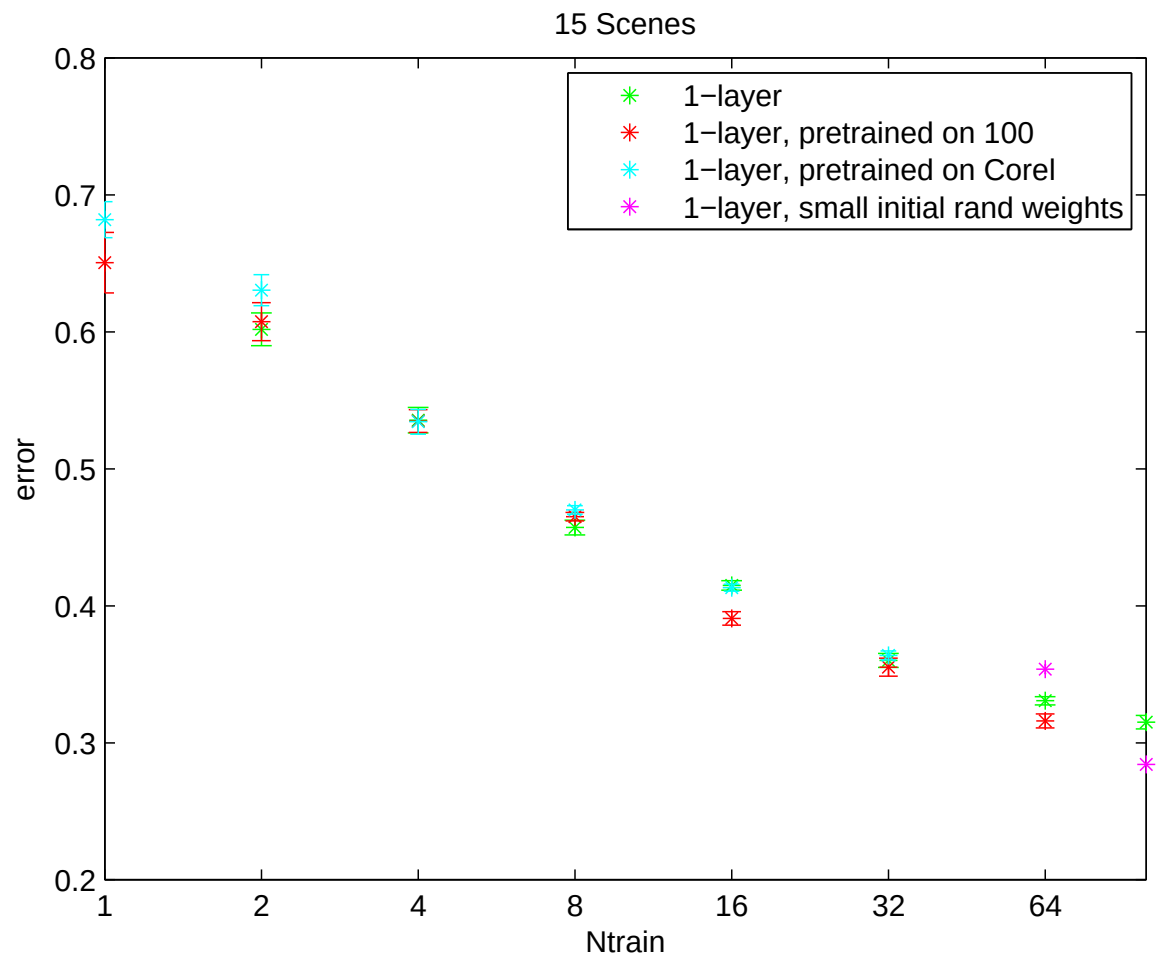

Figure 7: Comparison of different methods for one hidden layer system on 200 words. In the first set the system is pre-trained on the same set as used for back-prop, in the second it was pre-trained on 100 images per category, in the third on Corel database and in the fourth there was no pre-training (note: The last case for sizes 64 and 100 was obtained after 500 and 5000 iterations resp.)

Fig. 8. The one hidden layer system does the best, but that adding more hidden layers worsens the performance.

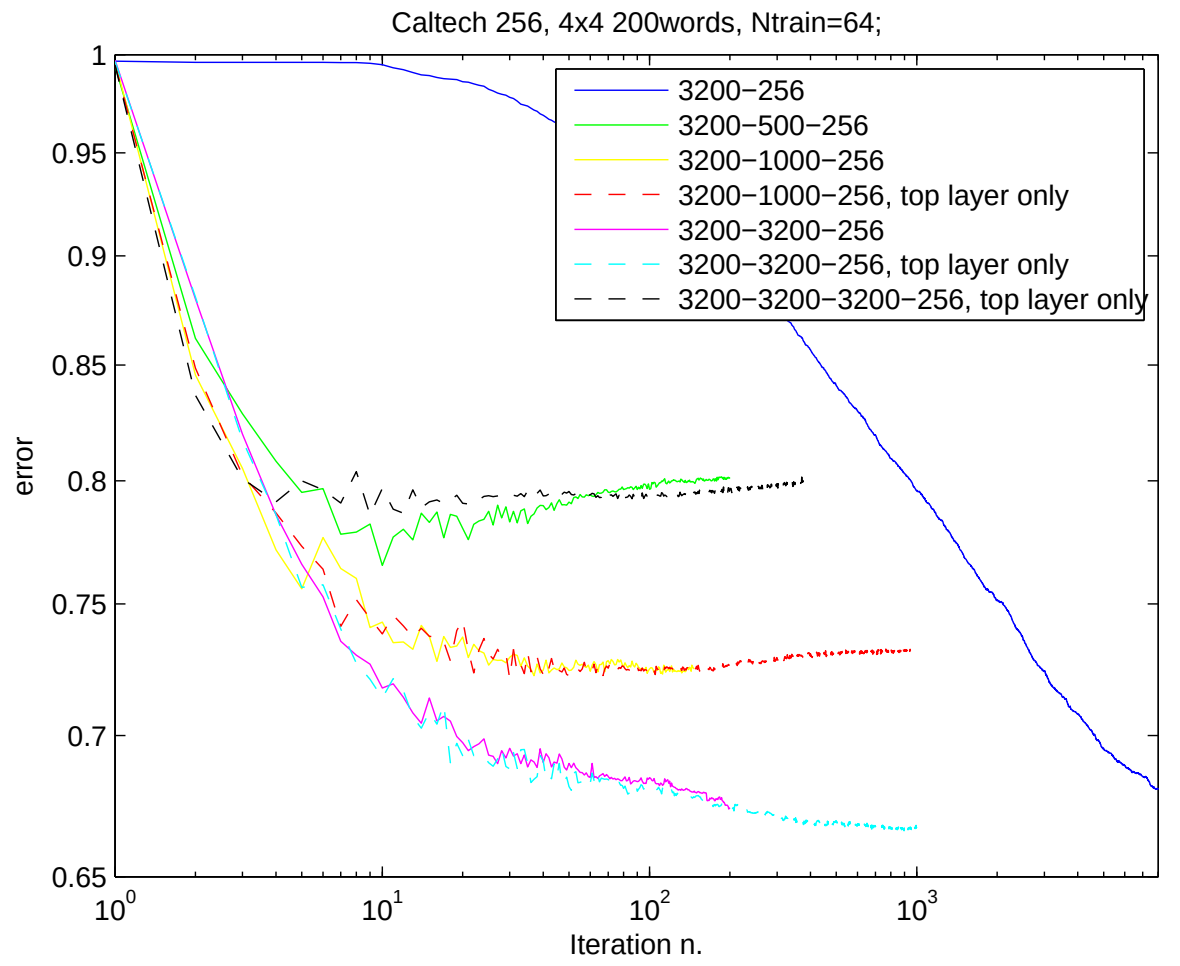

Figure 8: The performance on Caltech 256 bag of words obtained by breaking up every image into 4x4 pieces and treating same words in different regions as different.

Let us discuss the running times, starting with 1x1 case, selecting the appropriate numbers. The 3000 iterations of zero layer system took about 32 hours. 400 pre-training iterations for one layer system about 2 hours and 300 back-propagation periods about 15 hours. For the 4x4 case, the 8000 iterations of the zero layer system took 24 days. 200 pre-training iterations for one layer system took 21 hours and 300 back propagation periods of the top layer took about 29 hours.

Finally we compare the performance to the other methods. This is shown in the Fig. 9. We see that for bag of words the svm gives about the same results, may be slightly better then ours. For the 4x4 case, our result is not very good. However we didn't spend much time improving or analyzing the performance in this case. The graph also shows the much more impressive result by Varma (this is not applied to just the simple bag of these words that we use).

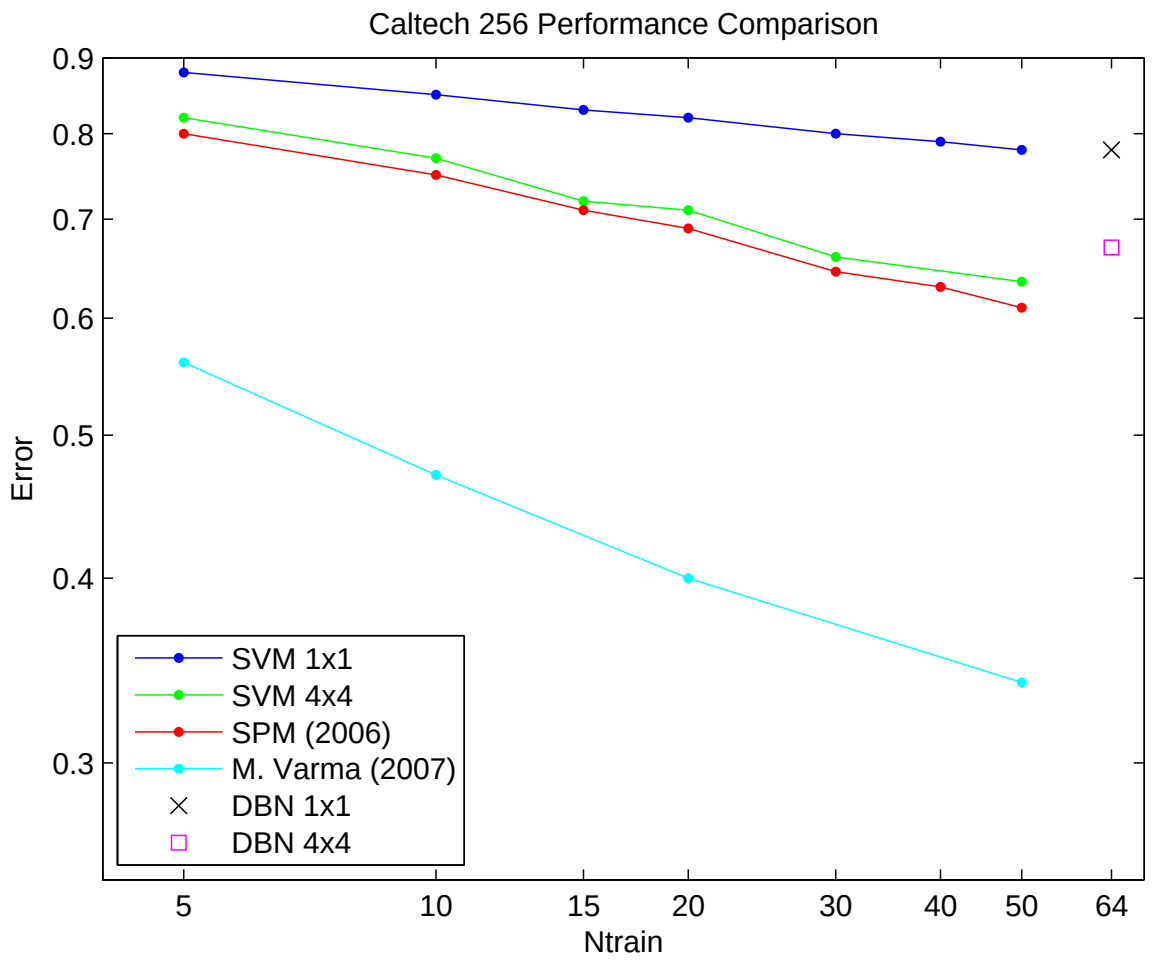

Figure 9: Comparison of performance for different methods. For notation see caption of Fig. 3.

# 5 Conclusions

Unsupervised learning discovers structure in images. This is strongly suggested by our experimental observations of a short supervised training phase, advantage of having a lot of unsupervised data, ability to learn from different dataset then the one with labels and in approximate emergence of neurons explicitly representing certain categories. However for these data, the effects are rather weak. Adding hidden layers doesn't improve the performance. This is probably due to the fact that histograms of quantized sift features either don't have very high level structure (which doesn't sound surprising) or not the kind of structure these networks can discover. This suggests that further research should be devoted to developing richer representations of image content, of a form that may be processed by an RBM-like network. Finally, as a classifier, the DBN's do quite well, but in this case slightly worse then the standard methods of SVM or SPM. Since SVM is pushed into perfection among shallow architectures, this should not be discouraging and one should focus on the problems where high level structure is needed.

We thank Pietro Perona and Yoshua Bengio for useful discussions, Yevgeni Bart for the 13 Scenes data and Caltech computer vision cluster for computing resources.